\documentclass[journal]{IEEEtran}
\usepackage{latexsym}

\usepackage{enumerate}
\usepackage{graphicx}
\usepackage{subfigure}
\usepackage{multirow}

\usepackage{framed,multirow}
\usepackage{times}
\usepackage{graphicx} % more modern
\usepackage{subfigure}
\usepackage{amssymb}
\usepackage{latexsym}

\usepackage{algorithm}
\usepackage{algorithmic}

\usepackage{hyperref}

\usepackage{url}
\usepackage{xcolor}
\usepackage{amsmath}
\usepackage{amsxtra}
\usepackage{braket}
\usepackage{array}
\usepackage{color}
\usepackage{url}
\usepackage{placeins}

\usepackage{times}
\usepackage{balance}
\usepackage{enumerate}
\usepackage{multirow}
\usepackage{pifont}
\usepackage{marvosym}
\usepackage{ifsym}
\usepackage{threeparttable}

\usepackage{lipsum} % 输出随机文本
\usepackage{amsmath,esint} % 输出数学公式
\usepackage{cuted}

\begin{document}
\title{SSHPool: The Separated Subgraph-based Hierarchical Pooling}
\author{Zhuo~Xu,~\IEEEmembership{}
        Lixin~Cui,~\IEEEmembership{IEEE~Member}
        Ming~Li,~\IEEEmembership{IEEE~Member}
        Yue~Wang,~\IEEEmembership{}
        Ziyu~Lyu,~\IEEEmembership{}
        Hangyuan~Du,~\IEEEmembership{}\\
        Lu~Bai${}^{*}$,~\IEEEmembership{IEEE~Member}
        Philip S.~Yu,~\IEEEmembership{IEEE~Fellow}
        Edwin R.~Hancock,~\IEEEmembership{IEEE~Fellow}

\thanks{

Zhuo Xu and Lu Bai (${}^{*}$Corresponding Author: bailu@bnu.edu.cn) are with School of Artificial Intelligence, Beijing Normal University, Beijing, China. Lixin Cui and Yue Wang are with School of Information, Central University of Finance and Economics, Beijing, China. Ming Li is with the Key Laboratory of Intelligent Education Technology and Application of Zhejiang Province, Zhejiang Normal University, Jinhua, China. Ziyu Lyu is with School of Cyber Science and Technology, Sun Yat-Sen University, Guangzhou, China. Hangyuan Du is with School of Computer and Information Technology, Shanxi University, Taiyuan, China. Philip S. Yu is with Department of Computer Science, University of Illinois at Chicago, US. Edwin R. Hancock is with Department of Computer Science, University of York, York, UK}% <-this % stops a space
}
\markboth{}%
{Shell \MakeLowercase{\textit{et al.}}: Bare Demo of IEEEtran.cls for Journals}

\maketitle

\begin{abstract}
In this paper, we develop a novel local graph pooling method, namely the Separated Subgraph-based Hierarchical Pooling (SSHPool), for graph classification. We commence by assigning the nodes of a sample graph into different clusters, resulting in a family of separated subgraphs. We individually employ the local graph convolution units as the local structure to further compress each subgraph into a coarsened node, transforming the original graph into a coarsened graph. Since these subgraphs are separated by different clusters and the structural information cannot be propagated between them, the local convolution operation can significantly avoid the over-smoothing problem caused by message passing through edges in most existing Graph Neural Networks (GNNs). By hierarchically performing the proposed procedures on the resulting coarsened graph, the proposed SSHPool can effectively extract the hierarchical global features of the original graph structure, encapsulating rich intrinsic structural characteristics. Furthermore, we develop an end-to-end GNN framework associated with the SSHPool module for graph classification. Experimental results demonstrate the superior performance of the proposed model on real-world datasets.
\end{abstract}

\begin{IEEEkeywords}
Deep Learning; Graph Neural Networks; Graph Pooling
\end{IEEEkeywords}

% make the title area
\maketitle
\IEEEpeerreviewmaketitle

\section{Introduction}\label{Sec1}
In recent years, deep learning methods such as the Convolutional Neural Network (CNN) have experienced a rapid developing process. Many effective CNN-based models have been applied to multiple research fields, including image segmentation~\cite{image_segmentation}, image classification~\cite{image_classification}, natural language processing~\cite{NLP}, etc. The great success of the CNN encourages researchers to further explore employing it on graph structure data~\cite{graph_data_1}, that usually has a complex structure and is a typical instance of irregular data in non-Euclidean space. Unlike the regular data such as images and vectors, the graph data requires more effective models, that can not only analyze the attributes of the elements in the graph structure but also figure out the structural relationships between the elements. Thus, it is difficult to directly employ the traditional CNNs for analyzing the graph data.

\subsection{Literature Review}\label{Sec1.1}
To effectively deal with the graph data, some researchers have tried to enhance the ability of CNN models for irregular grid structures of graphs. Meanwhile, the Graph Neural Networks (GNNs)~\cite{GNN} are proposed to accomplish graph-based tasks. By integrating the CNN and the GNN methods together, the Graph Convolutional Network (GCN)~\cite{GCN} has been rapidly developed, and its related approaches~\cite{GraphSAGE,DGCNN} have achieved outstanding performance in capturing the effective features of the graph data. For example, the Deep Graph Convolution Neural Network (DGCNN)~\cite{DGCNN} has leveraged the graph convolution layers with the SortPooling module to learn semantic characteristics. Based on the DGCNN models, some researchers further explore its application for different downstream tasks~\cite{DGCNN_app_1,DGCNN_app_2,DGCNN_app_3}. To accomplish the classification task, the Hierarchical Graph Convolutional Network (HGCN)~\cite{HGCN} has adopted the hierarchical graph convolution to aggregate structurally similar nodes, exhibiting its strong empirical performance on some specific downstream tasks (e.g., node classification). Furthermore, by associating with the attention mechanism, some derivative GNN models, such as the Graph Transformer Network (GTN)~\cite{GTN} and the Spatio-Temporal Graph Transformer Network (STGTN)~\cite{STAR}, have been developed and further improved the capability of extracting complicated graph features. These extended GNN-based models provide powerful tools for graph data analysis. However, when facing the graph representation learning task, there is usually additional requirements for these aforementioned models, i.e., they need to abstract features from original graphs with less redundant information~\cite{minCUT}. To this end, the downsampled filtering method, namely the graph pooling operation, is highlighted.

\begin{figure*}
\centering
\subfigure[The global pooling]{\includegraphics[width=.76\columnwidth]{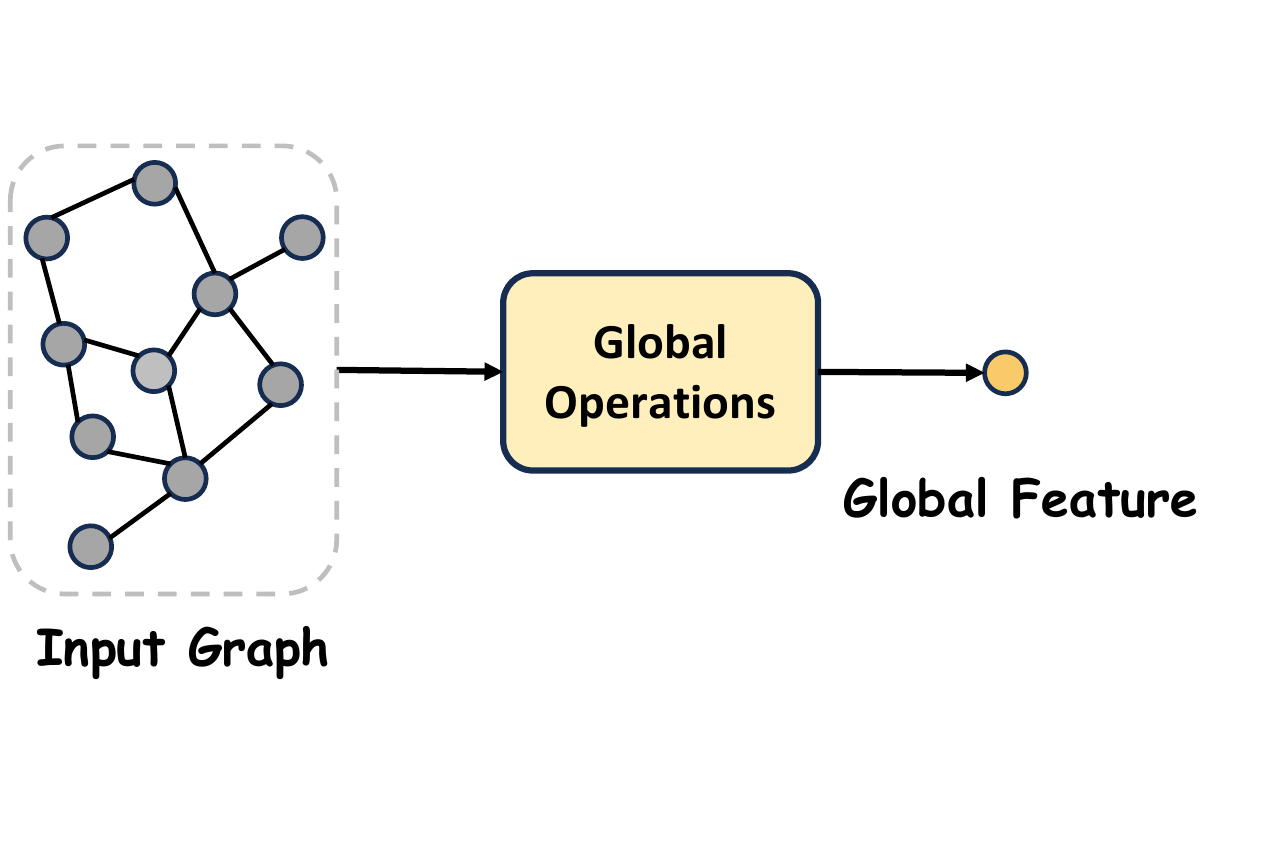}\label{global_pool}}
\subfigure[The local pooling]{\includegraphics[width=1.20\columnwidth]{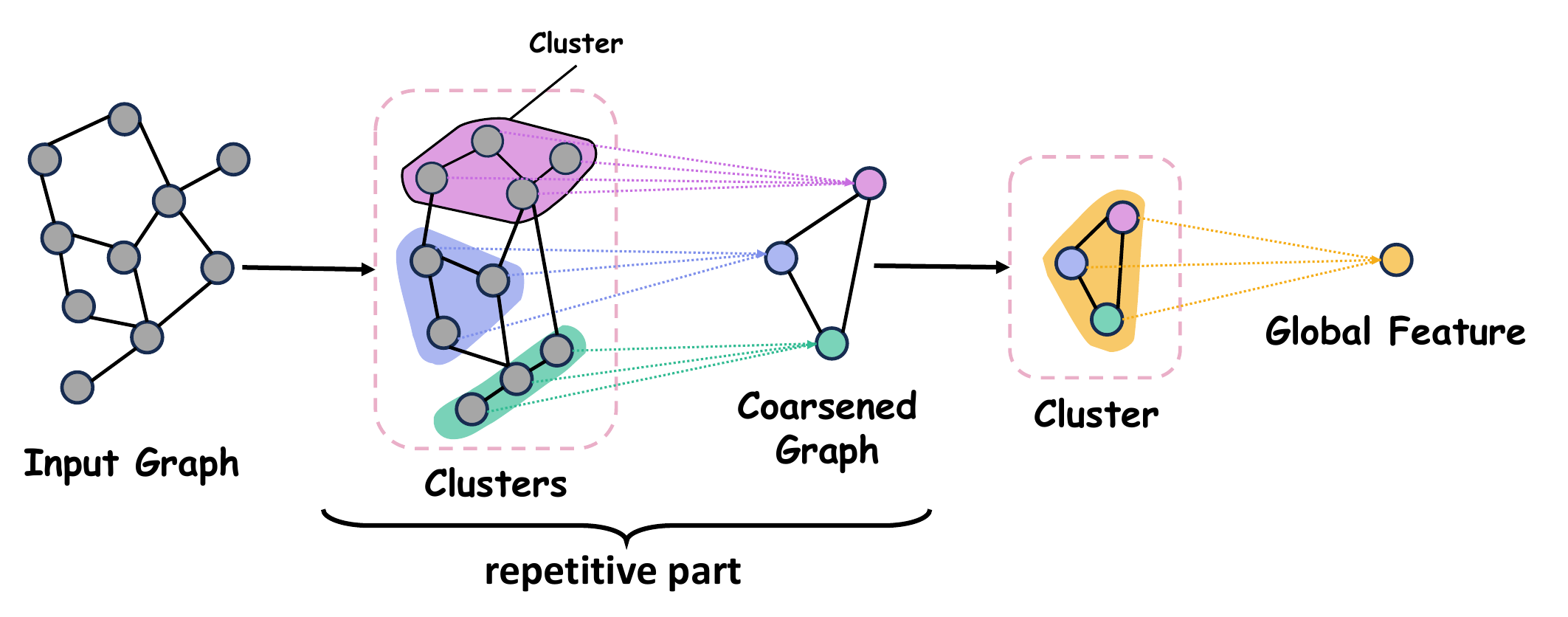}\label{local_pool}}
\vspace{-0pt}
\caption{Descriptions of two pooling methods. Both the global and the local pooling methods convert the original input graph into the global feature. On the left side, the global pooling tends to directly sum up or average the representations of all nodes over the whole graph structure. By contrast, on the right side, the local pooling tends to assigns the nodes into different clusters, and then hierarchically aggregates the nodes from the same cluster into a coarsened node, resulting in a family of layer-wise coarsened graphs.}
\label{pooling_methods}
\vspace{-0pt}
\end{figure*}

Similarly to the traditional pooling operation defined for the CNN models, the graph pooling operation proposed for the GNN models plays an important role in the sampling process. The main purpose of the graph pooling is to downsize the graph structure and make the representation learning efficient~\cite{purpose}. For existing works, the hierarchical scheme in the GNN models has significantly affected the definitions of novel graph pooling operations, and the researchers have focused more on the enhancement of the GNN models associated with the idea of hierarchical pooling~\cite{DiffPool,Asap,SUGAR,MTPool,ABDPooling}. Specifically, for the multi-layer framework of the GNN models, the original input graph can be gradually compressed into a family of coarsened graphs with shrinking sizes, so that the global feature of the original graph can be obtained through the coarsened graphs. According to the arrangement of nodes involved in the node assignment, there are two specific types of pooling methods for graphs, namely the global pooling and the local pooling. The descriptions of both the methods are shown in Figure~\ref{pooling_methods}.

Generally speaking, the global pooling is a common methodology in the pooling operation~\cite{global_pooling_01,global_pooling_02}. Global operations, such as the summing pooling and the averaging pooling, can directly compute the global representation of a graph by simply summing up or averaging all the node representations over the whole graph~\cite{global_pooling}. However, since the global pooling neglects the feature distribution associated with different nodes, it cannot identify the structural differences between the nodes. Thus, the global pooling can not effectively learn the topological information residing on the local nodes and may be unsuitable for a hierarchical structure. To overcome the shortcoming, the local pooling that focuses more on the features of local nodes is developed. With this strategy, a number of local pooling operations, namely the hierarchical pooling operations, based on hierarchical structures are developed. One typical approach of them is the DiffPool~\cite{DiffPool}. The core idea is to learn the layer-wise node assignment matrices, that assign the nodes to different clusters and then hierarchically aggregate the graph of the current layer into a coarsened graph for the next layer, by compressing the nodes belonging to the same cluster into a new coarsened node. However, the associated node assignment matrix is computed based on the soft assignment procedure, i.e., each node may be simultaneously assigned to multiple clusters associated with different probabilities. For the DiffPool, the receptive field of each local patch (i.e., the cluster) may influence that of other patches. Moreover, the coarsened node representation of the DiffPool is computed by directly summing up the node representations belonging to the same cluster, the DiffPool fails to identify the differences between different nodes.

Although the above hierarchical pooling methods have improved the performance of the GNN models, they still suffer from some problems when the graph is aggregated into a coarsened one. Specifically, we summarize the challenges arising in the existing methods as follows. \textbf{(a)} \textbf{Over-smoothing}~\cite{oversmoothing}. The node representation of each layer relies on the information propagation between neighbor nodes over the global coarsened graph structure. Thus, either the original nodes or the coarsened nodes can continuously propagate their information to all other nodes through the edges after multiple layers. As a result, the representations of different nodes tend to be similar or indistinguishable to each other. This phenomenon is the so-called notorious over-smoothing problem~\cite{Rethinking}, seriously influencing the performance of the GNN models. \textbf{(b)} \textbf{Degradation}. This issue arises in the context of the hierarchical pooling within GNNs, that adheres to deep learning architectures. When the GNN model associated with the pooling layers increases in depth, the degradation becomes a significant challenge, influencing the performance of the GNN model~\label{homogeneous}.
\subsection{Contributions}

The aim of our work is to overcome the aforementioned shortcomings by proposing a new pooling method, that focuses more on extracting local features from the specific separated subgraphs. One key innovation of the new pooling method is that we hierarchically decompose the original graph structure into the separated subgraphs, and extract the structural information from each individual subgraph. This scheme can in turn isolate the structural information between different subgraph structures, naturally providing an elegant way to reduce the over-smooth problem. Overall, the main contributions of this work are threefold.

\textbf{First}, we propose a novel local graph pooling method, namely the Separated Subgraph-based Hierarchical Pooling (\textbf{SSHPool}), to learn effective graph representations. We commence by assigning the nodes of each graph into different clusters, and this process results in a family of separated subgraphs where each subgraph consists of the nodes belonging to the same cluster and retains the edge connection between these nodes from the original graph. With these separated graphs to hand, we individually employ several local graph convolution units to compress each subgraph into a coarsened node, further transforming the original graph into a coarsened graph. Since each local convolution unit is restricted in each subgraph and the node information cannot be propagated between different subgraphs, our new pooling method can significantly reduce the over-smoothing problem and provide discriminative coarsened node representations for the resulting coarsened graphs. Furthermore, by hierarchically performing the proposed procedures on the resulting coarsened graph, the proposed SSHPool can effectively extract the hierarchical global feature of the original graph structure, encapsulating rich intrinsic structural characteristics.

\textbf{Second}, we develop an end-to-end GNN-based framework associated with the proposed SSHPool module for graph classification. Since the proposed SSHPool consists of hierarchical substructure sets, there are multiple layer-wise stacked local convolution unit sets associated with the hierarchical substructures, i.e., the SSHPool module may be deeper. To overcome the possible degradation problem that may be caused by the deeper hierarchical structures, we propose a graph attention layer for the proposed GNN-based model to accomplish the interaction between the global features extracted from the SSHPool and the initial embeddings from the original graph, further improving the effectiveness of the proposed GNN-based model.

\textbf{Third}, we empirically evaluate the classification performance of our GNN model associated with the proposed SSHPool module. Experimental results on standard graph datasets demonstrate the effectiveness.

This paper is organized as follows. Section~\ref{related_work} introduces the related work about our research. Section~\ref{methodology} presents the proposed approach in details. Section~\ref{experiments} expresses the performance of our model on real-world datasets, and analyzes the reason for the effectiveness. Section~\ref{conclusion} concludes this paper.

\section{Related Works}\label{related_work}
\subsection{The Graph Neural Networks}
GNNs have been widely employed for various graph-based applications~\cite{gnn_link_prediction,gnn_graph_classification,gnn_node_classification}. In these instances, the input data is represented as a graph structure, and the GNN aims to learn effective representations that can better describe the structural characteristics of graphs.

Specifically, given an input graph $G(V,E)$ with the node set $V$ and edge set $E$, assume $X \in \mathbb{R}^{n \times d}$ is the node feature matrix (i.e., each of the $n$ nodes has a $d$-dimensional feature vector) and $A$ is the adjacency matrix. The GNN model focuses on deriving the representations as follows, i.e,
\begin{equation}
    Z_{G} = \mathrm{GNN}(X,A;\Theta),
\end{equation}
where $Z_{G}$ is the extracted graph embedding matrix (i.e., the graph feature representation), and $\Theta$ is the set of training parameters for the GNN model. With these graph representations to hand, the downstream classifier can be employed to predict their corresponding labels.

Inspired by the traditional CNNs, the GNN methods have rapidly evolved and further formed the so-called Graph Convolution Networks (GCNs) for graph structures. For instance, the GCN model proposed by Kipf et al.,~\cite{GCN} implements the GNN model by combining the following layer-wise propagation scheme, i.e.,
\begin{equation}
    H^{(l+1)} =  \mathrm{ReLU}(\tilde{D}^{- \frac{1}{2}}\tilde{A}\tilde{D}^{- \frac{1}{2}}H^{(l)}W^{(l)}),\label{GCN_module}
\end{equation}
where $H^{(l)} \in \mathbb{R}^{n \times d}$ is the hidden embedding matrix from the $l$ layer, $W^{l} \in \mathbb{R}^{d \times d}$ is the trainable matrix for the $l$ layer, $\tilde{A} = A + I$ is the adjacency matrix associated with the self loop, $\tilde{D} = \sum_{j}\tilde{A}_{ij}$ is the corresponding degree matrix associated with $\tilde{A}$, and $H^{(l+1)}$ is the embedding matrix extracted for the next layer $l+1$ of the GNN model. Note that, $H^{(0)}$ corresponds to the node feature matrix $X$, i.e., the $H^{(0)} = X$ is the original graph feature representation. Eq.(\ref{GCN_module}) indicates that each module layer of the GNN model adopts the structural message propagation function to accomplish the representation learning. Unlike the traditional GNNs, the GCN is able to extract and aggregate the graph structure information within the neighborhood rooted at a node. With the increasing layers, the GCN can gradually encapsulate the high-order neighborhood information into the consideration.

\subsection{The Pooling Approaches}\label{Sec2.2}
As a kind of information filtering and downsizing  method, the pooling operations are widely used in the traditional CNNs. Compared to the large number of pooling strategies for the traditional CNNs, the amount of graph pooling methods is fewer. Generally speaking, there are two specific types of pooling methods for graphs, namely the global pooling and the local pooling. Specifically, the global pooling approaches tend to employ some simple operations to directly compute the global representations of graphs, such as summing up or averaging all the node representations over the whole graph~\cite{global_pooling}. Since the simple summing or averaging operation cannot identify the feature distribution of different nodes, the global pooling approaches usually have poor performance.

On the other hand, the local pooling approaches tend to gradually form a family coarsening graph with shrinking sizes to extract global features of graph structures, they can thus reflect rich local intrinsic structural information. As a typical instance of local pooling methods, the DiffPool~\cite{DiffPool} operation is proposed to learn hierarchical graph representations. Specifically, given the input graph adjacent matrix $A^{(l)} \in \mathbb{R}^{n_{l} \times n_{l}}$ and the node embedding matrix $X^{(l)} \in \mathbb{R}^{n_{l} \times d}$ for the layer $l$, the DiffPool layer coarsens the input graph as
\begin{equation}
    (A^{(l+1)}, X^{(l+1)}) = \mathrm{DiffPool}_{l}(A^{(l)}, X^{(l)}),
\end{equation}
where $A^{(l+1)}$ and $X^{(l+1)}$ are the corresponding adjacency matrix and  the node embedding matrix of the resulting coarsened graph for the next layer $l+1$, and $\mathrm{DiffPool}_{l}(\cdot)$ is the pooling function in the layer $l$. Since the DiffPool method can hierarchically learn node assignment matrices that  assign the nodes to different clusters, this pooling method can aggregate the graph of the current layer into a coarsened graph for the next layer by compressing the original nodes (i.e., $X^{(l)}$) belong to the same cluster into new coarsened nodes (i.e., $X^{(l+1)}$). Thus, the node number $n_{l+1}$ of $X^{(l+1)} $ is lower than that $n_{l}$ of $X^{(l)}$.

Based on the basic hierarchical scheme of the DiffPool method, a number of extended hierarchical pooling methods have been developed by redesigning the assignment procedure or the coarsening procedure~\cite{SAGPOOL,Asap,ABDPooling,C2NABDP,SEP,AdamGNN}. Specifically, these hierarchical pooling methods could be divided into the following categories, by considering whether the pooling operation needs to reserve all nodes.

\textbf{(a) Hierarchical pooling with partial nodes.} One typical idea is the Adaptive Structure Aware Pooling (ASAPool)~\cite{Asap}. After the assignment process, the ASAPool scores the different clusters in the graph. The top scored clusters will be selected and compressed into coarsened nodes. The other clusters which are considered as the redundant information are abandoned. As a result, the ASAPool operation relies on the partial nodes. Similar approaches also include the Subgraph Neural Network with the Reinforcement Pooling (SUGAR)~\cite{SUGAR} that samples the subgraphs from the original graph, and adopts the top $k$ strategies to select the necessary subgraphs, dropping partial nodes during the pooling process. \textbf{(b) Hierarchical pooling with all nodes.} One typical example is the attention-based differentiable pooling (ABDPool)~\cite{DiffPool}. Unlike the ASAPool, it tends to obtain the structural information from all node clusters.

\textbf{Remarks:} In summary, the pooling operation with partial nodes focuses on capturing the structural information from the key targets (e.g. node clusters, subgraphs), while the pooling operation with all nodes tends to encapsulate the structural information over the whole graph structure. Although these pooling methods significantly improve the existing GNN models, most existing hierarchical pooling still suffer from two drawbacks, that is the over-smoothing problem as well as the degradation problem. The first problem is due to the fact that the node representation of each hierarchical layer relies on the information propagation between neighbor nodes. The information of each node will be propagated to all other nodes through the paths between the nodes after multiple layers, resulting in similar or indistinctive node representations for all nodes. The second problem is due to the fact that these hierarchical pooling operations are usually associated with GNNs that typically have deep computational architectures, significantly resulting in degradation. Both problems limit the performance of the GNN-based models.

\section{The Methodology}\label{methodology}

To overcome the theoretical drawbacks of the existing hierarchical pooling approaches, in this section, we propose a novel Separated Subgraph-based Hierarchical Pooling (\textbf{SSHPool}) to learn effective graph representations. Similar to the other hierarchical pooling methods, our proposed approach can be applied to the GNN models. We first give the definition of the proposed SSHPool. Moreover, we develop an end-to-end GNN-based framework associated with the proposed SSHPool module for graph classification. Finally, we analyze the theoretical properties of the proposed SSHPool.

\subsection{Definitions of the Proposed SSHPool}\label{secpool_module}

\begin{figure}
    \centering
    \includegraphics[width=.47\textwidth]{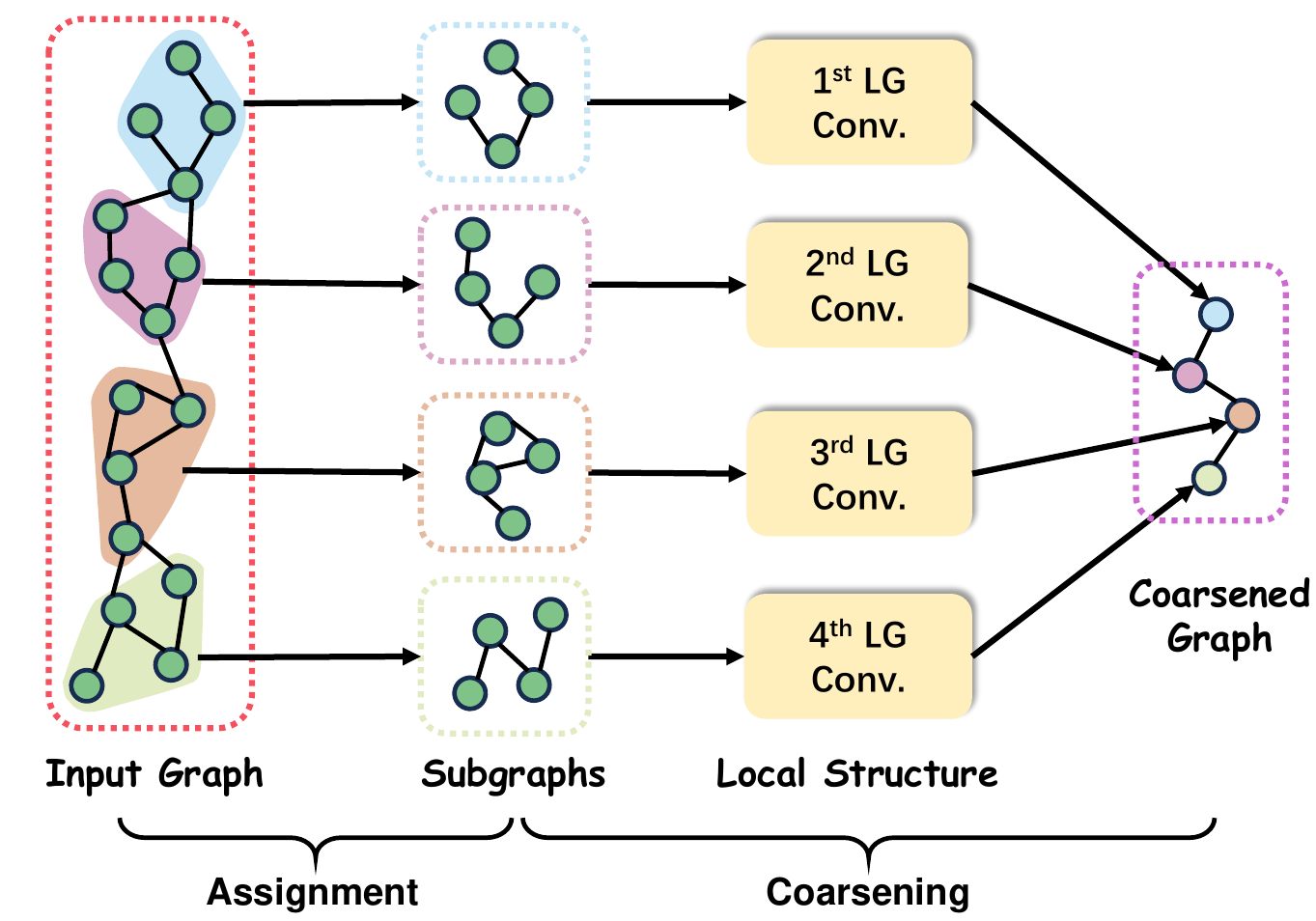}
    \caption{The computational architecture for the proposed SSHPool of each layer. The first step is the assignment where the input graph is decomposed into multiple separated subgraphs. Then, the second one is the coarsening process where these subgraphs are compressed into the nodes of the coarsened graph.}
    \label{secpool_overview}
\end{figure}

In this subsection, we propose a novel pooling method SSHPool, that is defined based on hierarchical separated subgraphs to reduce the over-smoothing problem arising in most existing hierarchical pooling methods. The overview of our proposed SSHPool for each computational layer is shown in Figure~\ref{secpool_overview}, and the computational architecture of \emph{\textbf{the proposed SSHPool for each layer}} mainly consists of two steps. \textbf{First}, given an input graph, the nodes are adaptively assigned into different clusters represented by different colors. This process in turn generates a family of separated subgraphs, where each subgraph consists of the nodes belonging to the same cluster and retains the edge connection between these nodes from the original graph. \textbf{Second}, to extract the local structural information of each subgraph, we individually employ several Local Graph Convolution (\textbf{LG Conv.}) units to propagate the node information between each individual subgraph and then aggregate the subgraph into a coarsened node. Note that, the parameters for the local convolution operations of different subgraphs are not shared. Since each local convolution unit is restricted in each subgraph and the node information cannot be propagated between different subgraphs, significantly reducing the over-smoothing problem and providing discriminative coarsened node representations for the resulting coarsened graphs, i.e., the input graph for the next layer of SSHPool. By stacking the above computational steps (i.e., the coarsened graph will be employed as the new input graph for the next layer), the proposed SSHPool can hierarchically extract the intrinsic graph characteristics. Below, we give the detailed definitions for the proposed SSHPool of each layer.

\textbf{The Assignment and the Local Graph Convolution.} For each layer $l$ of the proposed SSHPool, assume the input graph is denoted as $G^{(l)}=(V^{(l)}, E^{(l)})$ and has $n_{l}$ nodes in $V^{(l)}$ (i.e., $n_{l} = |V^{(l)}|$). Note that, $G^{(l)}$ can be either the input graph embedded from the global convolution (for $l=1$) or the coarsened graph from the last layer $l-1$ of the SSHPool (for $l\geq 2$). The first step of the SSHPool is to decompose $G^{(l)}$ into a family of $n_{l+1}$ separated subgraphs, by assigning the $n_l$ nodes of $G^{(l)}$ into $n_{l+1}$ clusters through an node assignment matrix. And we define the assignment ratio as  $\alpha = \frac{n_{l+1}}{n_{l}}$. Unlike the classical hierarchical based DiffPool~\cite{DiffPool}, the proposed SSHPool is based on the hard assignment matrix, i.e., each node cannot be assigned to multiple clusters. In fact, the hard assignment matrix can be computed through the soft one. Specifically, given the node embedding matrix $X^{(l)} \in \mathbb{R}^{n_l \times d}$ of $G^{(l)}$, the soft assignment matrix $S_{\mathrm{soft}}^{(l)}$ is defined as
\begin{equation}
\label{classifier}
    S_{\mathrm{soft}}^{(l)} = \mathrm{softmax}(X^{(l)}),
\end{equation}
where $S_{\mathrm{soft}}^{(l)}\in \mathbb{R}^{n_l \times n_{l+1}}$ ($n_{l+1}< n_l$), and $\mathrm{softmax}(\cdot)$ is the softmax classifier. With $S_{\mathrm{soft}}^{(l)}$ to hand, the $(i,j)$-th entry of the hard assignment matrix $S^{(l)}\in \{0, 1\}^{n_l \times n_{l+1}}$ satisfies
\begin{equation}
\label{assignment_equation}
     S^{(l)}(i,j) = \left\{
	\begin{aligned}
		1 \ & \mathrm{if} \ S_{\mathrm{soft}}(i,j) = \mathop{\max}\limits_{\forall j \in n_{l+1}}[S_{\mathrm{soft}}(i,:)];\\
		0 \ & \mathrm{otherwise}.
	\end{aligned}
	\right.
\end{equation}
Clearly, each $i$-th row of the hard assignment matrix $S^{(l)}$ selects the maximum element as $1$ and the remaining elements as $0$, i.e., the $i$-th node is only assigned to the $j$-th cluster. With $S^{(l)}$ to hand, we can decompose the input graph $G^{(l)}$ into $n_{l+1}$ separated subgraphs, and each $j^{\mathrm{th}}$ subgraph is denoted as $G_{j}^{(l)} = (V_{j}^{(l)},E_{j}^{(l)})$, where the node set $V_{j}^{(l)}$ consists of the nodes belong to the $j$-th cluster and the edge set $E_{j}^{(l)}$ remains the original edge connections between the nodes in $V_{j}^{(l)}$ from the original input graph $G^{(l)}$.

With the associated node feature matrix $X_{j}^{(l)} \in \mathbb{R}^{|V_{j}^{(l)}| \times d}$ and the adjacent matrix $A_{j}^{(l)} \in \mathbb{R}^{|V_{j}^{(l)}| \times |V_{j}^{(l)}| }$ of each $j$-th separated subgraph $G_{j}^{(l)}$ to hand, we propose a local graph convolution unit to extract the local structural information as
\begin{equation}
    Z^{(l)}_{j} = \tilde{A}_{j}^{(l)}X_{j}^{(l)}W_{j}^{(l)},
\end{equation}
where $\tilde{A}_{j}^{(l)} = A_{j}^{(l)} + I$,  $W_{j}^{(l)}\in \mathbb{R}^{d \times d}$ is the trainable weight matrix of layer $l$, and $Z^{(l)}_{j}\in \mathbb{R}^{|V_{j}^{(l)}| \times d}$ is the resulting local structural representation of $G_{j}^{(l)}$.

\textbf{The Coarsened Graph Generation.} After several individual local graph convolution operations on the separated subgraphs within the same layer of the proposed SSHPool, we aggregate the local information of different subgraphs to further generate a coarsened graph, as the input graph $G^{(l+1)}$ for the next layer of the proposed SSHPool. To this end, we collect the embedding local structural matrices of all the subgraphs, and compress each subgraph into the coarsened node.

In order to map the subgraphs to the coarsened graph nodes, we define a mapping vector $s^{(l)}_{j}$ to compress each subgraph $G_{j}^{(l)}$ into a coarsened node, and $s^{(l)}_{j} = S^{(l)}[:,j]$ where we split the hard assignment matrix into vector. And $s^{(l)}_{j}$ represents the nodes belonging to $j$-th subgraph.  Note that, $s^{(l)}_{j}$ is a vector with the empty nodes which are not in the $j$-th subgraph. So we transfer $s^{(l)}_{j}$ into $\hat{s}^{(l)}_{j}$ where the empty nodes are removed and the rest of the nodes remain in order.  With the hard assignment matrix $S^{(l)}$ defined by Eq.(\ref{assignment_equation}) and the mapping vector $\hat{s}^{(l)}_{j}$ of each $G_{j}^{(l)}$ to hand, the node feature matrix $X^{(l+1)}\in \mathbb{R}^{n_{l+1} \times d}$ and the adjacency matrix $A^{(l+1)}\in \mathbb{R}^{n_{l+1} \times n_{l+1} }$ of the resulting coarsened graph $G^{(l+1)}$ are defined as follows, i.e.,
\begin{equation}
X^{(l+1)}=  \mathop{\|} \limits_{j=1}^{n_{l+1}} \hat{s}^{(l)\top}_{j} Z^{(l)}_{j}    ,\label{Xequation}
\end{equation}
\begin{equation}
 A^{(l+1)}=  S^{(l)\top}A^{(l)}S^{(l)},\label{A_equation}
\end{equation}
where $\mathop{\|}$ is a concatenation operation of $\hat{s}^{(l)\top}_{j} Z^{(l)}_{j}$s.

\textbf{The Global Feature Generation.} The above definitions give the computational procedures of the proposed SSHPool for each layer $l$. By hierarchically stacking the computational procedures with multiple layers, the initial graph $G^{(0)}$, the proposed SSHPool can hierarchically extract the final global feature $\textit{\textbf{X}}$ for $G^{(0)}$ in the last layer $L$ ($l\leq L$). Specifically, given the initial input graph $G^{(0)}$ associated with its node feature matrix $X^{(0)}$ and the adjacent matrix $A^{(0)}$, the whole process of the proposed SSHPool can be generalized as
\begin{equation}
    \textit{\textbf{X}} = \mathrm{SSHPool}_{L}(X^{(0)},A^{(0)};\Theta_\mathrm{{S}}),\label{final_SSHPool}
\end{equation}
where $\mathrm{SSHPool}_{L}(\cdot)$ is the proposed SSHPool function with $L$ layers, and $\Theta_{S}$ represents the parameter set of SSHPool.

\begin{figure*}
    \centering
    \includegraphics[width=0.90\textwidth]{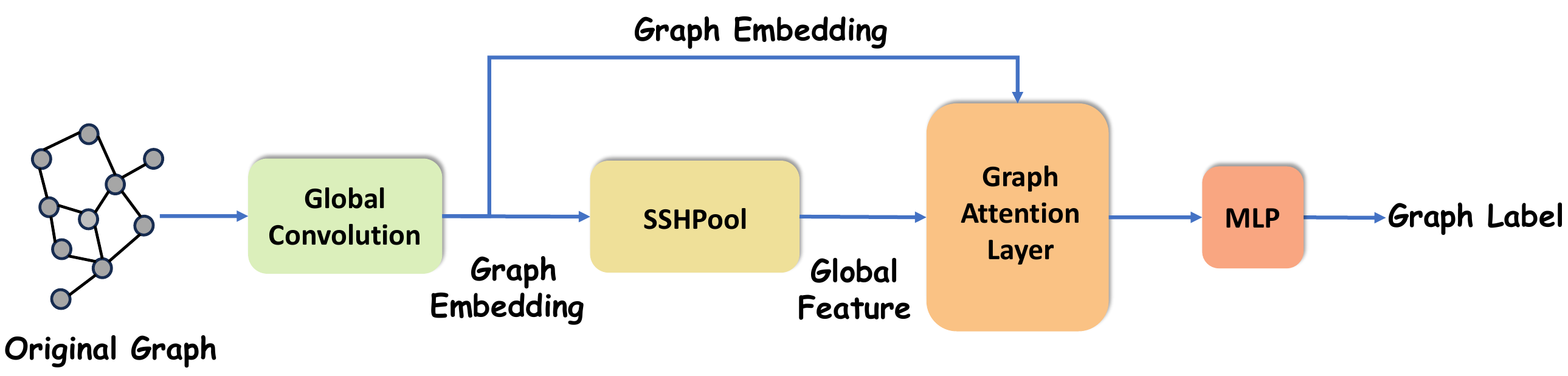}
    \caption{The computational architecture of our GNN framework.}
    \label{framework_overview}
\end{figure*}

\subsection{The GNN Framework associated with the Proposed SSHPool}\label{framework}

% , i.e., the proposed SSHPool has multiple stacked layers each of which is defined as the form of Figure 2.

Based on the proposed SSHPool method, we develop an end-to-end GNN framework for the graph classification. The computational architecture of our GNN framework is shown in Figure~\ref{framework_overview}, and has three main computational steps. First, for each original input graph, we commence by performing a standard global graph convolution operation~\cite{DGCNN} to generate the initial node feature matrix as the graph embedding. Second, with the graph embedding of each original graph to hand, we perform the proposed SSHPool to further extract the global hierarchical structural feature through multiple sets of layer-wise hierarchical separated subgraphs. Third, to overcome the possible degradation problem caused by the multiple hierarchical structure of the proposed SSHPool, we introduce an attention-based layer to further integrate the graph embedding and the global feature from the first and second steps, and the integrated graph feature is employed by the followed MLP for classification. Below we give the detailed definition of our attention-based layer.

\textbf{The Graph Attention Layer.} After the hierarchical pooling process of the SSHPool operation, the original graph feature has been transformed into the global feature in a hierarchical structure. Since there are multiple layer-wise stacked local convolution unit sets associated with the hierarchical substructures of the proposed SSHPool (i.e., the SSHPool module may be deeper), resulting in possible degradation problems. To resolve this problem, we propose a graph attention layer to further integrate the initial graph embedding extracted from the global graph convolution operation and the global feature extracted from the proposed SSHPool. Given the global feature $\textit{\textbf{X}}$ defined by Eq.(\ref{final_SSHPool}) and the  graph embedding $X^{(0)}$ from the global convolution operation, the function of the attention-based mechanism is able to reinforce the key graph feature and learn the cross-information, and attention calculation is defined as
\begin{equation}
    \begin{aligned}
    \mathcal{A} &= \mathrm{Attention}(X^{(0)}, \textit{\textbf{X}}) \\ &= \mathrm{softmax}\left(\frac{(\textit{\textbf{X}}W_{q})(X^{(0)}W_{k})^{\top}}{\sqrt{d}}\right)(X^{(0)}W_{v}),
    \label{graph_attention_layer}
    \end{aligned}
\end{equation}
where $W_{k},W_{q},W_{v} \in \mathbb{R}^{d \times d}$ are the corresponding parameter matrices, and $\mathcal{A}$ is the attention value. With the graph attention layer to hand, the proposed SSHPool can also encapsulate the initial graph embedding from the original graph structure input to our GNN framework, so that the parameters related to the input graph maintain their values and avoid the gradient vanishing caused by the degradation. Finally, we adopt the MLP as the classifier. Given the attention matrix $\mathcal{A}$ defined by Eq.(\ref{graph_attention_layer}) and the label set $\mathcal{Y}$, our GNN framework model computes the predicted graph labels as
\begin{equation}
    \hat{Pr} = \mathrm{Linear}(\mathcal{A}),
\end{equation}
where $\hat{Pr} = [\hat{p}_1,\hat{p}_2,...,\hat{p}_{\|\mathcal{Y}\|}]^{T}$ encapsulates the probabilities whose amount is $\|\mathcal{Y}\|$, and $\mathrm{Linear}(\cdot)$ is the active function of the MLP. And we adopt the cross entropy as the loss function in our training process.

\subsection{Discussions of the Proposed SSHPool}\label{discussion}

Compared to the aforementioned hierarchical-based graph pooling methods in Section~\ref{Sec1.1} and Section~\ref{Sec2.2}, the proposed SSHPool has some important properties, explaining the effectiveness. First, as we have stated in Section~\ref{Sec1.1} and Section~\ref{secpool_module}, unlike the classical hierarchical-based DiffPool, ABDPool, ASAPool, the proposed SSHPool is defined on the separated subgraphs, where there is no connection between the substructures. Thus, the proposed local graph convolution operation can only propagate the node structure information in each individual node cluster, i.e., the node information cannot be propagated between the nodes from different subgraphs. This can significantly reduce the over-smoothing problem and provide more discriminative structural characteristics for graphs. Since the proposed SSHPool focuses more on separated substructures, the resulting global graph features can encapsulating rich intrinsic structural information. Second, to overcome the possible degradation problem that maybe caused by the multiple layers of the proposed SSHPool, we also propose an attention layer for our GNN framework associated with the proposed SSHPool to simultaneously capture the initial graph embedding from the original graph structure, addressing the gradient vanishing caused by the degradation.

\section{Experiments}\label{experiments}

In this section, we evaluate the classification performance of our GNN-based framework associated with the proposed SSHPool on seven standard graph datasets~\footnote{http://graphlearning.io/}. Details of these datasets are shown in Table~\ref{dataset_description}.

\begin{table*}
\centering
% \scriptsize
\footnotesize
% \tiny
\caption{Dataset Description}\label{dataset_description}
%\vspace{-10pt}
\begin{tabular}{|c|c|c|c|c|c|}
\hline
\textbf{Datasets} & Vertices(max.)~ & Vertices(mean.)~ & Graphs & Vertex labels & Classes \\ \hline
D\&D                & 5748            & 284.32           & 1178   & 82            & 2       \\ \hline
PTC               & 64              & 14.29            & 344    & 18            & 2       \\ \hline
PROTEINS          & 620             & 39.06            & 1113   & 3             & 2       \\ \hline
NCI1              & 111             & 29.87            & 4110   & 37            & 2       \\ \hline
NCI109            & 111             & 29.68            & 4127   & 38            & 2       \\ \hline
IMDB-B            & 136             & 19.77            & 1000   & null          & 2       \\ \hline
IMDB-M            & 89              & 13.00            & 1500   & null          & 3       \\ \hline
\end{tabular}
%\vspace{-10pt}
\end{table*}

\subsection{Comparisons with Classical GNN and Pooling Methods}\label{settings}

\textbf{Experimental Settings:} We evaluate the classification performance of our GNN-based framework associated with the proposed SSHPool operation, and compare it against several alternative methods, including one classical graph kernel associated with C-Support Vector Machine (C-SVM)~\cite{cortes1995support}, seven GNN-based models with different pooling methods or optimization approaches. All these baseline models for the evaluation are introduced as follows.
\begin{itemize}

\item \textbf{WLSK}~\cite{GK} is one of the most popular graph kernel that is based on the subtree invariants.
\item \textbf{DGCNN}~\cite{DGCNN} adopts graph convolution model with the SortPooling layer that sorts graph nodes in a consistent order, resulting a standard grid structure for standard convolution operations.
\item \textbf{SUGAR}~\cite{SUGAR} is an embedding-based GNN model with hierarchical subgraph-level pooling.
\item \textbf{DiffPool}~\cite{DiffPool} is a GNN model with differentiable graph pooling module that generates representations of coarsened graphs. In the hierarchical architecture, there is an assignment matrix to compress the graph.
\item \textbf{ASAPool}~\cite{Asap} utilizes a modified GNN formulation and a self-attention network to focus on node information from the graph. It learns a sparse soft cluster assignment for nodes at each layer.
\item \textbf{ABDPool}~\cite{ABDPooling} leverages an attention-based differentiable pooling operation for learning a hard cluster assignment.
\item \textbf{SEPool}~\cite{SEP} proposes a hierarchical pooling approach with the GNN, and proposes a global optimization algorithm designed to generate the cluster assignment matrices for the pooling operation.
\item \textbf{AdamGNN}~\cite{AdamGNN} proposes a unified GNN-based model to interactively learn node and graph representations through a mutual-optimization method. This method focuses on multi-grained semantics of nodes.
\item \textbf{SSHPool(non)} is our GNN framework associated with the proposed SSHPool. But it does \textbf{not} include the Global Attention Layer.
\item \textbf{SSHPool} is our GNN framework associated with the proposed SSHPool.
\end{itemize}

For the alternative graph kernels, we search the optimal hyperparameters for the graph kernel on each dataset. For the GNN models or the GNN models associated with the proposed and alternative graph pooling operations, the Adam optimizer is set with the same learning rate of $1e-3$ in our experiments, and each method is trained with 100 epochs. Moreover, during the training process of these GNN models, we set the hidden dimension as 128, the dropout as 0.5, and the batch size as 32. In order to compare all these methods fairly, for either the graph kernels or the GNN models we adopt the 10-fold cross validation to compute the classification accuracy, and repeat the experiment for 10 times on each dataset. We show the average classification accuracy ($\pm$ standard error) and the average rank for each method in Table~\ref{experiment_result}. Finally, for the proposed SSHPool, we set the largest layer $L$ as $3$, then node numbers of the three layers follow $\{128, 32, 8\}$ where the assignment ratio $\alpha$ is 0.25.

% But the other baseline pooling methods and SSHPool(non)  are set to 2 layers, and the node number follows {128, 64}.

% The details of the discussion about pooling layer number are shown in the section~\ref{Sensitive}.

\begin{table*}[htbp]
\centering
\caption{Classification Accuracy (In $\pm$ Standard Error) for Comparisons.}\label{experiment_result}
% \scriptsize
\footnotesize
%\vspace{-10pt}
\begin{tabular}{|c|c|c|c|c|c|c|c|c|}
\hline
Datasets         & D\&D                  & PTC                 & PROTEINS            & NCI1                & NCI109              & IMDB-B              & IMDB-M              & Avg. Rank     \\ \hline
WLSK             & 74.38$\pm$0.69          & 59.23$\pm$0.45          & 71.70$\pm$0.67          & 70.32$\pm$0.40          & 69.71$\pm$0.52          & 64.48$\pm$0.90          & 43.38$\pm$0.75          & 9.15          \\ \hline
DGCNN            & 75.34$\pm$0.78          & 58.34$\pm$2.39          & 73.21$\pm$0.34          & 67.78$\pm$1.02          & 67.42$\pm$0.67          & 67.45$\pm$0.83          & 46.33$\pm$0.73          & 9.00          \\ \hline
SUGAR            & 78.62$\pm$0.26          & 65.72$\pm$2.15          & 76.81$\pm$0.31          & 74.93$\pm$1.34          & 73.72$\pm$0.66          & 71.43$\pm$0.93          & 48.12$\pm$1.76          & 3.86          \\ \hline
DIFFPool         & 77.24$\pm$0.46          & 63.39$\pm$1.03          & 74.86$\pm$0.35          & 62.32$\pm$1.90          & 61.98$\pm$1.98          & 70.12$\pm$0.63          & 47.20$\pm$1.81          & 8.14          \\ \hline
ASAPool          & 75.75$\pm$0.45          & 64.71$\pm$2.41          & 75.53$\pm$0.94          & 71.48$\pm$0.42          & 70.07$\pm$0.55          & 71.25$\pm$0.58          & 48.67$\pm$1.34          & 5.86          \\ \hline
ABDPool          & 74.13$\pm$0.52          & 63.82$\pm$2.37          & 73.24$\pm$0.91          & 71.54$\pm$1.28          & 71.78$\pm$1.35          & 70.58$\pm$0.71          & 50.63$\pm$1.47          & 6.43          \\ \hline
SEPool           & 77.43$\pm$0.23          & 65.34$\pm$1.03          & 75.32$\pm$0.34          & 73.78$\pm$0.31          & 73.17$\pm$0.42          & 70.34$\pm$0.53          & ~50.31$\pm$1.72         & 4.86          \\ \hline
AdamGNN          & 80.12$\pm$0.34          & 66.73$\pm$0.68          & 77.04$\pm$0.78          & 75.67$\pm$0.30          & 72.99$\pm$0.52          & 72.85$\pm$0.67          & 49.35$\pm$0.96          & 2.86          \\ \hline
SSHPool(non)     & 79.31$\pm$0.36          & 63.91$\pm$1.35          & 77.18$\pm$0.43          & 74.18$\pm$0.45          & 73.33$\pm$0.38          & 71.80$\pm$0.39          & 48.13$\pm$1.03          & 3.86          \\ \hline
\textbf{SSHPool} & \textbf{81.80$\pm$0.48} & \textbf{67.74$\pm$1.43} & \textbf{79.38$\pm$0.28} & \textbf{75.91$\pm$0.14} & \textbf{74.45$\pm$0.16} & \textbf{73.02$\pm$0.47} & \textbf{51.14$\pm$1.28} & \textbf{1.00}  \\ \hline
\end{tabular}
%\vspace{-10pt}
\end{table*}

\textbf{Experimental Results and Analysis:} Table~\ref{experiment_result} indicates that our GNN model associated with the proposed SSHPool can outperform all the alternative methods on the seven standard datasets. The effectiveness of our GNN model associated with the proposed SSHPool is threefold. \textbf{First}, all the alternative GNN models or the GNN models associated with different alternative graph pooling operations mainly rely on the node information propagation over the global graph structure. As a result, all these alternative methods suffer the over-smoothing problem, and the node representations extracted by these methods tend to be similar and not discriminative for classification. By contrast, our proposed SSHPool can adaptively decompose the input graph into separated subgraphs, and thus perform the individual local convolution operation on each subgraph. Since there is no connection between different separated subgraphs, the convolution operation can only propagate the node information within each subgraph and the node information cannot be propagated between different subgraphs. As a result, the proposed SSHPool can significantly reduce the over-smoothing problem and provide discriminative structural representations for graphs. \textbf{Second}, the WLSK kernel associated with a C-SVM is a typical instance of shallow learning. Moreover, the required kernel computation is also separated from the C-SVM classifier, and cannot provide an end-to-end learning mechanism. By contrast, our GNN model associated with the proposed SSHPool is an end-to-end deep learning approach, naturally having better classification performance. \textbf{Third}, the GNN models associated with the alternative pooling operations may suffer from the degradation problem, since the associated hierarchical pooling mechanisms usually have multiple layer-wise hierarchical structures. By contrast, to overcome the possible degradation problem that may be caused by the deeper hierarchical structures, we employ an attention layer to accomplish the interaction between the global features extracted from the SSHPool and the initial embeddings from the original graph, further improving the effectiveness of the proposed GNN-based model.

Finally, we observe that the proposed SSHPool-based GNN model significantly outperforms the SSHPool(non) based model, i.e., the proposed SSHPool without the Global Attention Layer. This indicates that our Global Attention Layer can really address the possible degradation problem caused by the deeper architecture of the proposed SSHPool. Moreover, although the SSHPool(non) cannot achieve the best performance due to the possible degradation problem, it still outperforms most of the alternative methods, especially the GNN models associated with other pooling methods in terms of the averaged ranking. This again indicates that the separated subgraphs associated with the local graph convolution operations can tremendously improve the effectiveness.

\subsection{The Sensitive Analysis}\label{Sensitive}

\begin{figure}[htbp]

    \centering
    \includegraphics[width=0.5\textwidth]{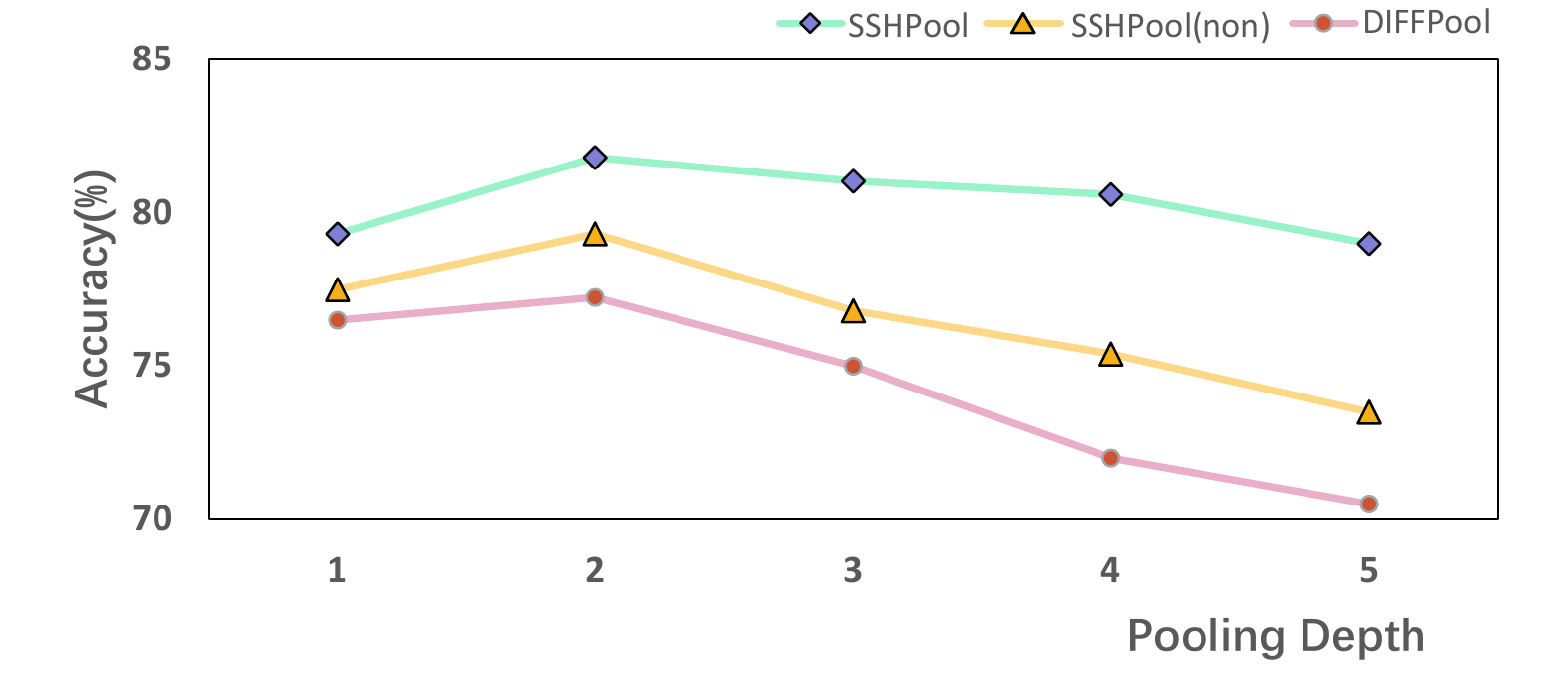}

    \caption{Analysis with the depth.}\label{Sen}

\end{figure}

\textbf{The Depth of the SSHPool.} To explore the effectiveness of the proposed SSHPool one step further, we evaluate how the classification performance is influenced by the depth of the SSHPool (i.e., the stacked layers of the SSHPool). Moreover, we also perform the same experiment with the DiffPool and the SSHPool(non). For the experiments, we utilize the D\&D dataset as an instance for these pooling methods, since we will observe the similar phenomenon on other datasets. Specifically, the experimental results are shown in Figure~\ref{Sen}. We observe that the proposed SSHPool can outperform the alternative pooling methods associated with all different pooling depths, again demonstrating the effectiveness of the proposed SSHPool. Moreover, we observe that the accuracies tend to be lower with the deeper depth, due to the possible degradation problem. However, we find that the accuracies of the proposed SSHPool tend to decrease slowly than the alternative pooling methods, again demonstrating the Global Attention Layer for the proposed SSHPool can significantly address the degradation problem.

\begin{figure}[htbp]
    \centering
    \includegraphics[width=0.5\textwidth]{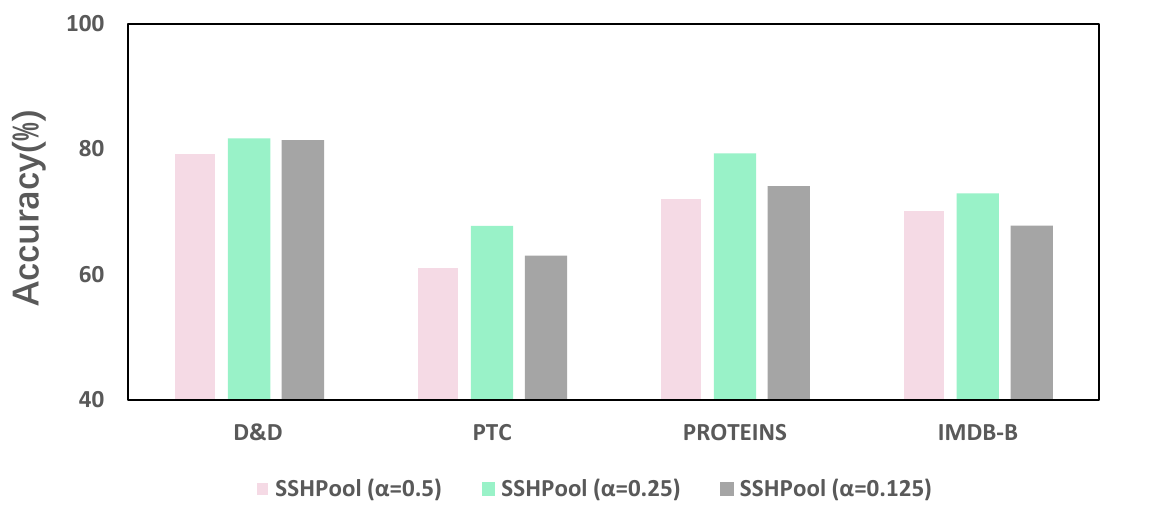}
    \caption{Analysis with the assignment ratio.}\label{Sen_node_assign}
\end{figure}

\textbf{The Assignment Ratio of the SSHPool.} To further explore the assignment process of the proposed SSHPool, we evaluate how the classification performance varies associated with different node assignment ratios as $\alpha$. Specifically, we set the ratios as $\alpha \in \{0.5, 0.25, 0.125\}$. We adopt the D\&D, PTC, PROTEINS, and IMDB-B for this evaluation and the experimental results are shown in Figure~\ref{Sen_node_assign}. Overall, the SSHPool associate with the assignment ratio $\alpha = 0.25$ outperforms that associated with the assignment ratios $\alpha = 0.125$ and $\alpha = 0.5$, i.e., the assignment ratios can significantly influence the performance of the proposed SSHPool. This is due to the fact that the assignment ratio determines the number of the separated subgraphs for the next layer of the SSHPool. If the ratio is very small, the nodes of the current layer will be assigned into less clusters, and the size of the resulting coarsened graph from the SSHPool will be quite small than the original graph. On the other hand, if the ratio is very large, the nodes of the current layer will be assigned into more clusters, and the size of the resulting coarsened graph from the SSHPool tends to be as same as the original graph. Both of the above situations cannot represent a fine hierarchical structural information for the SSHPool. This evaluation indicates that the appropriate selection of the assignment ratio is necessary for the proposed SSHPool.

\section{Conclusion}\label{conclusion}

In this paper, we have developed a novel SSHPool method, that cannot only extract effective hierarchical characteristics of graphs but also address the over-smoothing problem through the local separated substructures. Moreover, we have developed a GNN model associated with the proposed SSHPool for graph classification. Experimental results have demonstrated the effectiveness of the proposed SSHPool. Moreover, we also discuss the reasons why our proposed SSHPool is successful in addressing these problems.

%\section*{Acknowledgments}
%This work is supported by the National Natural Science Foundation of China under Grants T2122020, 61976235, U21A20473, 62172370 and 61602535. Corresponding Author: Lu Bai (bailu@bnu.edu.cn; bailucs@cufe.edu.cn).

% was supervised by Dr. Lu Bai and Dr. Lixin Cui for his M.sc degree and

%-------------------------------------------------------------------------

\balance

%-------------------------------------------------------------------------
%\nocite{ex1,ex2}
%\bibliographystyle{latex12}
%\bibliography{XBib}

\bibliographystyle{IEEEtran}
\bibliography{mybibfile}

\end{document}